\documentclass{article}

\usepackage{arxiv}

\usepackage[utf8]{inputenc} 
\usepackage[T1]{fontenc}    
\usepackage{hyperref}       
\usepackage{url}            
\usepackage{booktabs}       
\usepackage{amsfonts}       
\usepackage{nicefrac}       
\usepackage{microtype}      
\usepackage{lipsum}
\usepackage{graphicx}
\graphicspath{ {./images/} }

\title{A Deep Q-Learning/Genetic Algorithms based novel methodology for optimizing COVID-19 pandemic government actions}

\author{
 Luis Miralles-Pechu\'an \\
  CeADAR (Ireland’s centre for Applied AI) \\ University College Dublin\\
  Dublin, Ireland\\
  \texttt{luis.miralles@ucd.ie} \\
   \And
 Fernando Jim\'enez \\
  Department of Information and \\ Communication Engineering. \\ University of Murcia.\\
  Murcia, Spain\\
  \texttt{fernan@um.es} \\
  \And
 Hiram Ponce \\
  Facultad de Ingeniería.\\
  Universidad Panamericana\\
  Mexico City 03920. Mexico \\
  \texttt{hponce@up.edu.mx} \\
  \And
 Lourdes Martínez-Villaseñor \\
  Facultad de Ingeniería.\\
  Universidad Panamericana\\
  Mexico City 03920. Mexico \\
  \texttt{lmartine@up.edu.mx} \\
}

\begin{document}
\maketitle
\begin{abstract}
Whenever countries are threatened by a pandemic, as is the case with the COVID-19 virus, governments should take the right actions to safeguard public health as well as to mitigate the negative effects on the economy. In this regard, there are two completely different approaches governments can take: a restrictive one, in which drastic measures such as self-isolation can seriously damage the economy, and a more liberal one, where more relaxed restrictions may put at risk a high percentage of the population. The optimal approach could be somewhere in between, and, in order to make the right decisions, it is necessary to accurately estimate the future effects of taking one or other measures.

In this paper, we use the SEIR epidemiological model (Susceptible - Exposed - Infected - Recovered) for infectious diseases to represent the evolution of the virus COVID-19 over time in the population. To optimize the best sequences of actions governments can take, we propose a methodology with two approaches, one based on Deep Q-Learning and another one based on Genetic Algorithms. The sequences of actions (confinement, self-isolation, two-meter distance or not taking restrictions) are evaluated according to a reward system focused on meeting two objectives: firstly, getting few people infected so that hospitals are not overwhelmed with critical patients, and secondly, avoiding taking drastic measures for too long which can potentially cause serious damage to the economy. We created three different scenarios and we defined a reward system to prioritize the sequences of actions that fulfilled the objectives.

The conducted experiments prove that our methodology is a valid tool to discover actions governments can take to reduce the negative effects of a pandemic in both senses. We also prove that the approach based on Deep Q-Learning overcomes the one based on Genetic Algorithms for optimizing the sequences of actions.
\end{abstract}


\section{Introduction}
It is quite clear that 2020 will be remembered as the year in which the world was hit by Severe Acute Respiratory Syndrome Coronavirus 2 (SARS-CoV-2) virus responsible for COVID-19 disease. The virus has generated devastating effects on public health and on the economy of many countries around the globe \cite{Covidpandemic}. What started in China as a potential threat to the whole world, has taken many countries by storm and the virus galloped over the population from one country to another. Unlike the seasonal flu, the coronavirus is a new strain to which the population has no immunity, and that explains why it spreads so quickly. In addition to that, according to the WHO, COVID-19 is around ten times more lethal than the typical flu \cite{lai2020severe}. 

To avoid the spread of the virus, the authorities of most countries applied severe restrictions such as lock-downs, confinements, or closing borders. Those restrictions had devastating effects on the economy \cite{fernandes2020economic}. Around April 2020, in just a few weeks, most of the non-essential businesses that require people's physical presence such as restaurants, hotels, cinemas, universities closed down. A domino effect took place and a high percentage of the population lost their job, which generated a very difficult scenario for the government of each country. There was some uncertainty about when the situation will come back to normal. The COVID-19 pandemic became overnight a game-changer in the economy and in the lifestyle of billions of people. It is hard to remember a moment in which the world was hit so quickly in a similar way \cite{fernandes2020economic}.

In a crisis situation, it is crucial for governments taking actions to protect the economy and public health of their countries. For example, not saturating the healthcare system capacity is of paramount importance. If the number of cases exceeds the number of available beds or ventilators at hospitals, the mortality rate can rapidly increase, provoking a tragic situation \cite{he2020coronavirus}. Broadly speaking, there are basically two extreme approaches of the spectrum in a situation like this. On the one hand, a restrictive one, which is protecting the population by imposing the use of global self-isolation. And on the other hand, a more relaxed approach which consists in not taking any action at all. The first one has a much higher negative effect on the economy of a country since most businesses have to close down whereas the latter favours much more the economic flow but can provoke a high number of deaths. There are some other approaches in between such as partial lock-downs, encouraging the population to keep the social distance, facilitating the use of masks, and encouraging frequent washing of hands \cite{world2020rational}.

It is very important for the government to anticipate the pandemic by taking the best actions in every situation, and to that end, it is of great help to create mathematical models to represent the behaviour of infectious diseases. The SEIR (Susceptible, Exposed, Infectious, Recovered) model has been used for decades and it is a recommended model for infectious diseases that have an incubation period \cite{wearing2005appropriate}. Once we have a model able to simulate the behaviour of the pandemic over the population, we can simulate the evolution of the virus for a given action. By doing so, we can answer questions like, for example, What is going to happen if we take action A or B? or, What percentage of people will be infected after three weeks of taking action A? We can also estimate how the taken actions will affect the economy of the country by the number of people that will lose their job or the number of companies that will have to close down \cite{barro2020coronavirus}.

To find out the optimal sequences of actions for a given country, first, we have to define a reward system to evaluate how good a sequence of actions is. For example, if by following a given actions the number of people requiring ICU (Intensive care units) is higher than the available ones, then we give to that sequence of actions a penalization (or negative reward). Similarly, if the economy suffers and people lose jobs, we also give a penalization. 

The presented problem is very similar to any Markov Decision Processes (MDP) where there are different states and there is an agent that takes different actions and gets rewards from the environment for those actions \cite{puterman2014markov}. The problem consists of solving which is the optimal policy, i.e., to take the best action for each state. There is a Reinforcement Learning algorithm called Q-Learning highly recommendable for finding the optimal policy. In particular, due to the high number of states in coronavirus simulations, we implemented Deep Q-Learning (DQL). DQL is a version of Q-Learning, very suitable for simulations, that approximates the Q-Table with an Artificial Neural Network \cite{hester2018deep}. The optimal policy is the one that takes actions over time and gets the highest accumulative reward. Additionally, we also implemented an approach based on Genetic Algorithms (GA) \cite{whitley1994genetic}, where individual chromosomes represent the sequences of possible actions. 

The implemented SEIR model is relatively simple because our contribution in this research is neither to make a perfect model to represent the COVID-19 pandemic nor to generate an accurate model for any specific country. Our main contribution is to propose a methodology based on DQL and GA to facilitate governments taking the best actions for an epidemiological pandemic. To our knowledge it has not being presented yet a paper on that specific topic before.

This document has been structured as follows. In section II we present the state of the art of SEIR models, GA, DQL and some of the literature about how Artificial Intelligence (AI) has been applied to mitigate the COVID-19 pandemic. In section III, we describe the proposed methodology for finding out the optimal sequence of actions a government can take depending on objectives. In section IV three different experiments are conducted to prove that the presented methodology is feasible and also to compare the performance of the approach based on DQL against the approach based on GA. In section 5, we present the conclusions and some future lines of work.

\section{State of the art}
In this section, we describe the state of the art of the essential components of our methodology: The SEIR model that simulates the spread of the COVID-19 \cite{d2002stability} in the population, and two other techniques implemented to discover the best actions for combating the pandemic according to the goals of each government, DQL \cite{wang2016does} and GA \cite{whitley1994genetic}. Additionally, we describe some of the latest studies about how AI has been used to combat the COVID-19 pandemic, with special emphasis on those investigations related to Deep Reinforcement Learning.

\subsection{SEIR epidemiological model for COVID-19}{\label{SEIR-Model}}
Mathematical models for infectious diseases are a suitable resource for understanding and controlling the spread of a virus in a population. These models use a series of variables such as the incubation period ($\alpha$), the average infectious days ($\gamma$), and the contact rate ($\beta$), modelled by a series of differential equations to represent the behaviour of a pandemic on a population. These mathematical models are developed from the hypothesis of case distribution for infectious diseases based on the Kermack and McKendrick theory in 1927 \cite{kermack1927contribution}, and are a generic model for any infectious disease. Each particular disease has a different coefficient for the variables ($\alpha,\beta,$ and  $\gamma$), but the equations to simulate its behaviour are the same for representing all diseases \cite{li1995global}.

\begin{figure}[ht]
\centering
\includegraphics[width=0.7\linewidth]{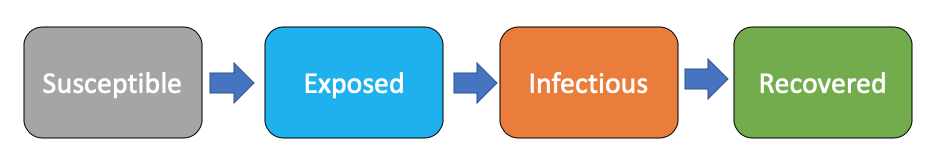}
\caption{Population groups to represent the SEIR mathematical modelling for infectious diseases.}
\label{fig:Deseases-flow}
\end{figure}

The SEIR model is a version of the original SIR model \cite{kermack1927contribution}. The SIR model divided the population into three categories: Susceptible (\textit{S}) that represent the healthy individuals, Infectious (\textit{I}) representing those individuals that transmit the disease and lastly, the Recovered (\textit{R}) those individuals that have recovered from the disease or have died from it. Due to the incubation period, a new category called Exposed (E) emerged representing those individuals that have contracted the disease but are still not able to transmit it \cite{d2002stability}. As can be seen in Figure \ref{fig:Deseases-flow}, most individuals start in group \textit{S}, then they move to group \textit{E} when they contract the disease. After the incubation period, they become part of the next group, \textit{I}, in which they are able to transmit it, and finally, after the transmission period, they move to the recovered group \textit{R}, either because they overcome the disease or because they died of it and are not able to transmit it \cite{li1995global}.

We use the following ordinary differential equations to represent the behaviour of the models which are modelled with the following variables:

\begin{itemize}
\item $\alpha$ is the inverse of the average incubation period in days.
\item $\beta$ or contact rate represents the average number of people an individual is in touch with and therefore can potentially transmit the disease.
\item $\gamma$ is the inverse of the average period in days in which an individual is infectious.
\end{itemize}

To model the behaviour of the disease, many simplified models consider the population Equation (\ref{eq:number1}), by which during the pandemic the population $N$ will remain constant. There are no changes in population numbers affected by people coming/leaving the country or by demographics.
\begin{equation}\label{eq:number1}
  N = S + E + I + R
\end{equation}
\textit{S} represents the susceptible group (healthy population). In the beginning, the vast majority of the population will be in this group. As in Equation (\ref{eq:number2}), the $S$ group will diminish over time as the Infectious group, $I$, is in contact with susceptible population.
\begin{equation}\label{eq:number2}
 \frac{dS}{dt} = - \beta SI
\end{equation}
The group of \textit{E} (Exposed) represents people infected who are not yet infectious due to the incubation period. It is modeled by Equation (\ref{eq:number3}). The \textit{E} population grows as the \textit{S} group gets infected and is reduced when the incubation period finishes. Those individuals go to the\textit{ I }(Infectious) group.
\begin{equation}\label{eq:number3}
  \frac{dE}{dt} = \beta SI - \alpha E
\end{equation}

As shown in Equation (\ref{eq:number4}), the group able to infect other people grows as the incubation period of the \textit{E} group finishes and diminishes as the average infections time $\gamma$ finishes.

\begin{equation}\label{eq:number4}
  \frac{dI}{dt} = \alpha E - \gamma I
\end{equation}

The last group is formed by the individuals from the \textit{I} group that change their state either because they recovered or because they died. As shown in Equation (\ref{eq:number5}), the sooner people recover, the shorter the infectious period. Since in this representation we are not considering the option of recovered individuals getting infected again, most individuals will end up in group \textit{R} as the days go by.

\begin{equation}\label{eq:number5}
\frac{dR}{dt} =  \gamma I
\end{equation}

\begin{figure}[t]
\centering
\includegraphics[width=0.50\linewidth]{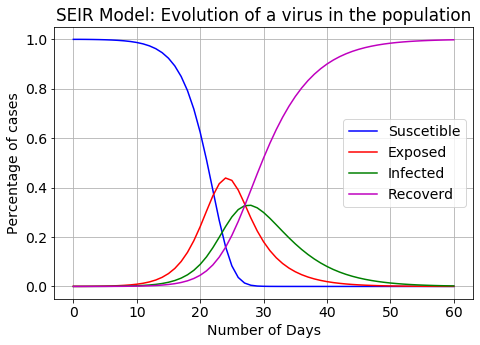}
\caption{Evolution of the infected and exposed group over a period of time. Simulation example over 60 days with $\alpha = 0.2$, $\beta = 2$ and $\gamma = 0.2$ in a population size of $N=50$M from from which 0.05\% are Infectious.}
\label{fig:simulation}
\end{figure}

In Figure \ref{fig:simulation} can be seen a simulation of a general infectious disease over 60 days with $\alpha$ = 0.2, $\beta$ = 2, and $\gamma = 0.2$. The total population was 50M and the initial infected was 10,000. It can be seen that by the end of the 60 days, almost all the population has been infected. That is why it is so important to take restrictive actions in advance in order to avoid oversaturation in the public health that implies a high fatality rate over the population.

\subsection{Deep Q-Learning}
Reinforcement Learning is one of the areas in artificial intelligence that has expanded most in recent years. Partly because of its great achievement of mastering the game of Go and beating the world champion Lee Sedol by 4-1 \cite{wang2016does}. In RL, an agent learns the right actions for each state of the environment from the rewards it obtains. The optimal policy is the one that maximizes the accumulative rewards of the agent from its initial state to the final state. The algorithm to learn which are the best actions is called Q-Learning. 

The Q-Learning algorithm is based on estimating the value for each pair of state-action. In other words, given the experience, it tells the agent in each state which action will provide higher accumulative rewards in the future. Q-Learning utilizes a table to store the estimated value for each action-state pair. called Q-Table. When Q-Learning is applied to environments with hundreds of different states, the Q-Table is replaced with an Artificial Neural Network (ANN) that approximates this table.

The interesting part of this algorithm is that it is able to learn from scratch by trial and error. There are some interesting publications that show how DQL was able to learn to play seven different Atari Games at a superhuman level \cite{mnih2013playing} learning by itself. DQL is particularly suitable for discovering which are the best sequences of actions across the states to get to the higher accumulative rewards since it is able to link one state with the others, thanks to the Bellman's equation \cite{hester2018deep}. Once trained, the agent is able to accurately estimate the reward summation from the current state to the final state.

\subsection{Genetic Algorithms}

Genetic algorithms (GA) were developed by John Holland in the 1960s \cite{whitley1994genetic}. GAs are well-known in computer science for optimizing problems of very different natures based on the famous theory of natural evolution and adaption by Charles R. Darwin. Such theory says that the fittest individuals are the ones with more chances of surviving and reproducing. The new elements will inherit the genes of their parents. Therefore, by this mechanism, the individuals of new generations will have higher chances of surviving and will be fitter than the previous ones. A population is formed by individuals, where each individual is represented with a chromosome and a chromosome is formed by a sequence of genes.

\subsection{Artificial Intelligence to Fight COVID-19}

Within the short period of 2020, AI has been used to analyse big data of COVID-19 patients to fight the disease \cite{alimadadi2020artificial,naude2020artificial}. According to the early review of Naude \cite{naude2020artificial}, AI can contribute in six areas to fight against COVID-19: (i) early warnings and alerts, (ii) tracking and prediction, (iii) data dashboards, (iv) diagnosis and prognosis, (v) treatment and cures, and (vi) social control. In this subsection we review the work in some of these areas focusing on social control which relates the most to our work. 

As described in \cite{alimadadi2020artificial}, machine learning techniques are used in taxonomic classification of COVID-19 genomes \cite{randhawa2020machine}, COVID-19 detection assay \cite{metsky2020crispr}, survival prediction of COVID-19 patients \cite{yan2020prediction}, and discovering potential drugs against this disease \cite{ge2020data}.

Deep Learning techniques are currently been used through image classification for COVID-19 detection and diagnosis \cite{wang2020covid,li2020artificial,gozes2020rapid,rao2020identification}. Support Vector Machines have also been used for COVID-19 detection \cite{barstugan2020coronavirus}.

Similarly, AI forecasting methods have been used to predict COVID-19 cases \cite{hu2020artificial}. For example, Al-qaness et al. \cite{al2020optimization} presented an improved adaptive neuro-fuzzy inference system to estimate and forecast the number of confirmed cases of COVID-19 in the upcoming ten days in China.  Ardabili et al. \cite{ardabili2020covid} presented a comparative analysis of machine learning (multi-layered perceptron and adaptive neuro fuzzy inference system)  and evolutionary models (genetic algorithms, particle swarm optimizer and grey wolf optimizer) to predict the COVID-19 outbreak. Their results show that multi-layered perceptron is an effective tool to model the outbreak.
Jahanbin and Rahmanian \cite{jahanbin2020using} approach is interesting given that they study unstructured data from Twitter and then analysed how to send an alert message to surveillance systems for timely detection outbreaks of COVID-19. The authors used a fuzzy rule-based evolutionary algorithm called Eclass1-MIMO.

Hoertel et al. \cite{hoertel2020facing} performed a stochastic agent-based microsimulation model of the COVID-19 epidemic in New York City and evaluated the potential impact of different decisions namely quarantine duration, type of quarantine lifting, post-quarantine screening and the use of hypothetical effective treatment against COVID-19. The impact was simulated on the disease's incidence and mortality, and on ICU-bed occupancy. 

Yang et al. \cite{yang2020modified} analysed the  policies adopted in China on January 2020 given the outbreak of COVID-19 originated in Wuhan. They considered the large-scale quarantine control, strict travel controls and extensive monitoring of suspected cases. They implemented a modified SEIR epidemiological model that incorporates domestic migration data and COVID-19 epidemiological data in order to assess the effects of these strategies and predict the epidemic progression. They used a Long-Short-Term-Memory model to predict the new infection peak.

\subsubsection{Reinforcement learning for epidemic control}

In order to develop prevention strategies to fight epidemics of infectious diseases, it is important to model the complex epidemic dynamics. These models enable governments to make predictions and study the effect of different strategies.

DQL can help to find optimal intervention policies to control an epidemic. Libin et al. \cite{libin2020deep} developed a DQL approach to learn prevention strategies in the context of pandemic influenza.  They aimed to model the effectiveness of school closure intervention using the SEIR model in 379 administrative regions in Great Britain. They demonstrated that DQL can be useful in the context of epidemiological decision-making. Khadilkar et al. \cite{khadilkar2020optimising} proposed a methodology to compute optimal lockdown/release policies learnt with a RL algorithm in the context of COVID-19 pandemic. They considered the function of disease parameters (infectiousness, gestation period, duration of symptoms, and probability of death) and population characteristics (density, movement propensity). Their policy includes health and economic costs.

A microscopic multi-agent epidemic model considering the consequences of an individual’s decision on the spread of COVID-19 was introduced by Liu \cite{liu2020microscopic}.  In this model, every agent can choose between staying healthy by limiting activities and maintaining high activity levels for living to minimize its cost function. Game theory and multi-agent reinforcement learning were used to optimize the individual agent decisions in order to predict the spread of the disease. They pointed out that other public policies must be taken into account in future work. Our contribution, although similar to that of Liu \cite{liu2020microscopic} and Libin et al. \cite{libin2020deep}, is more focused on facilitating governments setting goals and giving them the sequences actions (where each actions can be of a different confinement level) that satisfy those goals.

\section{Description of the proposed methodology}

In this section, we describe a methodology that can be used by different governments to optimize the best actions. Our proposed methodology is easily adaptable to any other country by changing the coefficient of the SEIR model, by adding more actions, or by changing the rewards and penalization to match the governments’ objectives. It is also open to incorporate more complex SEIR models. The proposed methodology uses a SEIR model to represent an infectious disease over the population and an optimization module which can be implemented with any optimization algorithm. In this paper, we are comparing GA against DQL. 

\subsection{Optimization model formulation}{\label{model-formulation}}

In order to solve the SEIR problem with GA and DQL, we need to find out the sequences of actions per day, where each action can be one of four possible ones (hard confinement, soft confinement, two-meter distance, and no restrictions). We defined a reward system that prioritises the sequences of actions that do not jeopardise the economy and that safeguard the healthcare system (although there is an implicit trade-off between the two objectives). 

We can formulate the problem in terms of mathematical programming as follows:

\begin{equation}
\label{optimization-model}
Maximize \ f(\textbf{x})=
\displaystyle\sum_{i=1}^n X_i(\textbf{x})
\end{equation}

\noindent with:
$$
X_i(\textbf{x})=\left\{
\begin{array}{ll}
RW_i & I_i\leq N_{beds} \\
P & otherwise \\
\end{array}
\right.
$$

\noindent where $\textbf{x}=(A_i)_{i=1}^n$ is a sequence of $n$ actions (one action for each of the $n$ days), with $A_i\in\{1,\ldots,M\}$, and $M$ is the number of possible confinement levels. $RW_i\in \mathbb{N}$ is a reward expressed in economic terms associated to each action $A_i$, $N_{beds}$ is the number of beds available, and $P$ is a penalty that represents a negative reward. $I_i$  is the number of infectious patients on  a day $i$, $i=1,\ldots,n$, and $I_0$ represents the initial number of infectious patients. The same applies to the Susceptible ($S_i$), Exposed ($E_i$), and Recovered ($R_i$) groups with initial conditions $S_0$, $E_0$ and $R_0$, respectively. According to Equations (\ref{eq:op_S}) to (\ref{eq:op_R}), the number of people for each group ($S_i$, $E_i$, $I_i$, and $R_i$) on each day $i$, $i=1,\ldots,n$, are calculated, from the initial values $I_0$, $R_0$, $S_0$ and $E_0$, as follows:

\begin{equation}\label{eq:op_S}
S_i= S_{i-1} - \delta_{i}\beta S_{i-1} I_{i-1}
\end{equation}

\begin{equation}\label{eq:op_E}
E_i= E_{i-1} + (\delta_{i}\beta S_{i-1} I_{i-1} -\alpha E_{i-1})
\end{equation}

\begin{equation}\label{eq:op_I}
I_i= I_{i-1} + (\alpha E_{i-1}-\gamma I_{i-1})
\end{equation}

\begin{equation}\label{eq:op_R}
R_i= R_{i-1} + (\gamma I_{i-1})
\end{equation}

\noindent where $\delta_i=[0,1]$, $i=1,\ldots,n$ is a weight associated with the contact rate $\beta$ for the $A_i$ action. When restrictive actions are applied the contact rate diminishes and we model that multiplying $\beta$ by $\delta$, where $\delta$ has a small value for restrictive confinement levels. Once the optimization problem has been defined, we have implemented the methodology with DQL and GA to see which one works better. Both methodologies address the same problem but each from its own perspective.

\subsection{Deep Q-Learning approach}

DQL is a well-known algorithm for solving RL-based problems \cite{hausknecht2015deep}. In our particular case, we have an initial state in which there are four groups. There are four actions the agent can take: apply total confinement, soft confinement, some restrictions, or not taking any action. Each action has a different impact on the contact rate and also on the economy (e.g. the more it reduces the contact rate the more it undermines the economy).

\begin{figure}[t]
\centering
\includegraphics[width=0.70\linewidth]{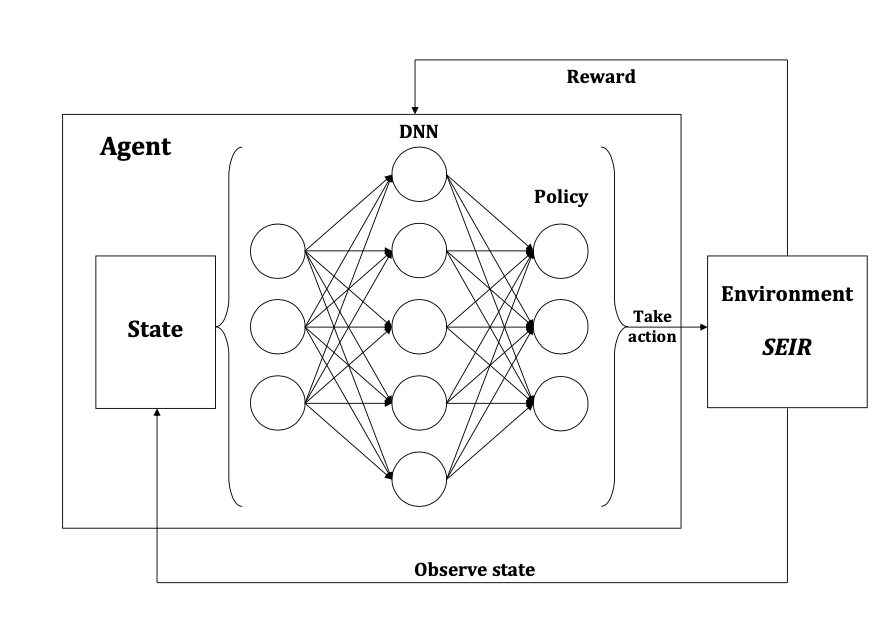}
\caption{Representation of a Markov Decision Process in which an agent tries to find out the optimal sequence of actions by interacting with the environment.}
\label{fig:DQL-Agent}
\end{figure}

The proposed methodology for DQL can be understood as a classic MDP problem represented in Figure \ref{fig:DQL-Agent} and is based on evaluating the proposed sequences of actions by giving a series of rewards. For example, every time the number of beds is exceeded a big penalization is given. Similarly, when a lock-down is applied, it implies a lower reward than when there are no restrictions. In such a way that the agent tries to discover those action sequences where the ICUs of the health system are not saturated, and the negative impact on the economy is minimized. From an RL perspective, the first day is the initial state of the agent, and the last day, to which we want to take actions, is the final state. The agent has to find out the optimal policy using the Q-Learning algorithm, which tells the agent which action to take in each state to drives it to maximize the accumulative rewards \cite{watkins1992q}.

\subsection{Genetic Algorithm approach}

For the GA approach, an individual $\textbf{x}=(A_i)_{i=1}^n$ is represented as an array of size $n$ where each element of the array is an integer number $A_i\in\{1,\ldots,M\}$, as shown in Figure \ref{fig:individual}. We used $M=4$ in the experiment ($M$ confinement levels). At the beginning, an initial population formed by chromosomes is randomly generated. We can see the sequence of actions as the chromosome of the individual and each of the actions as a gene. A gene represents the phase (level of confinement) a government can take for a given day.

\begin{figure}[t]
\centering
\includegraphics[width=0.5\linewidth]{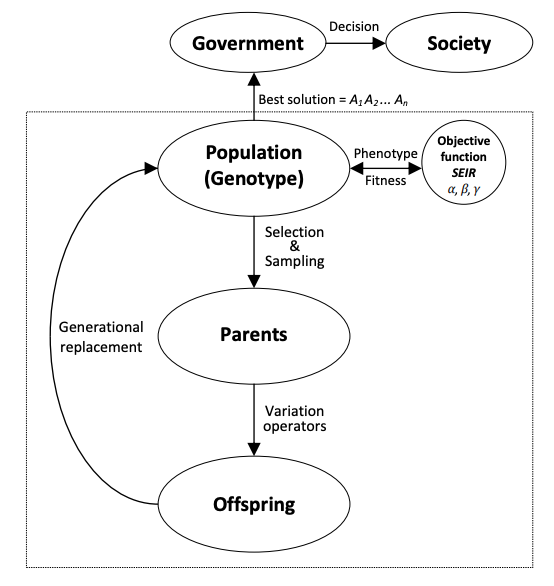}
\caption{GA representation for optimizing the sequences of actions (phenotype) a government can take in a pandemic.}
\label{fig:MethodologyGA}
\end{figure}

Figure \ref{fig:MethodologyGA} represents the methodology implemented with GA where the fitness function is the total cumulative of rewards for a specific individual. The individuals with better fitness will be the parents of the next generations. The individuals will cross between themselves and mutate to generate new individuals representing better solutions. Finally, the best sequence is given to the government so that it can apply it to society.

\begin{figure}
    \centering
    $$
\underbrace{\begin{array}{|l|l|l|l|l|}\hline
A_1 & A_2 & A_3 & \ldots & A_n \\\hline
\end{array}}_{Individual \ \textbf{x}}
\label{fig:individual}
$$
    \caption{Encoding of the individual in GA.}
    \label{fig:individual}
\end{figure}

\section{Experiments and results}

In this section, first, we explain the parameter configuration for conducting the experiments. Subsequently, we describe the configuration for the SEIR model, the DQL model and the GA. Then, we consider three different scenarios in which we prove if our methodology is a valid tool for optimizing sequences of actions in a pandemic. The results of the experiments representing the performance of two approaches of our methodology are displayed in different graphics and tables. Lastly, the results of the experiments are analyzed and discussed.

\subsection{Description of the experiment}\label{problem}

We define the phases a government can take during a time interval expressed in days from more restrictive to less restrictive. Remember that $\delta$ is the variable that reduces the contact rate and the more restrictive the action, the more the contact rate is reduced. We also have set that each day in a normal situation 10 billion are generated and that when restrictions are applied this quantity diminishes.

Each action for each day belongs to one of the following phases:

\begin{enumerate}
\item \textbf{$Phase{_1}$}: Total isolation, in which $\delta$ is 0.25 and the economy is reduced to 40\%. 
\item \textbf{$Phase{_2}$}: Partial isolation, in which $\delta$ is 0.50 and the economy is reduced to 60\%. 
\item \textbf{$Phase{_3}$}: Soft restrictions, simply maintaining two-meter distance and washing hands. In this scenario, $\delta$ is 0.75 and the economy is reduced to 80\%. 
\item \textbf{$Phase{_4}$}: No restrictions, in which the contact rate does not shrink, that is to say, $\delta$ is 1, but the economy is not affected (economy still at 100\%).
\end{enumerate}

The objective for the experiment to find the sequence of actions that has the lowest negative effect on the economy as long as the number of people does not exceed the maximum number of beds. To guarantee that there are enough beds during the pandemic, we give a big penalty of -1000 (an equivalent of what is generated in 100 days) for each day the number of patients requiring ICU exceeds the number of beds. The model has to find the right action among four different phases for 200 days. That is, $4 ^ {200}$ (or 2.58225e+120) combinations. The higher the total rewards, the better is the sequence since our rewards are aligned with the objective of the government. If the government had different goals such as eradicating the pandemic at any cost, then we just have to change the reward system.

\subsection{Parameter configuration}
For conducting the experiments we apply three different approaches: DQL, GA, and Random. Random consists of selecting the best performance of 1000 sequences of actions randomly generated. The parameters for the three modules, SEIR model, GAs, and DQL have been set as follows:

\subsubsection{SEIR model configuration}

There are some values such as the contact rate of the population that depends on the country where the pandemic takes place. For this particular experiment we have set the following values:
\begin{itemize}
\item $\beta$ representing the contact rate has been set to 4. $\alpha$ has been set to $0.1923$ because according to \cite{chen2020mathematical}, the Incubation Period is 5.2 days. Likewise, $\gamma$ has been set to $0.1724$ since according to \cite{chen2020mathematical} the infectious period for COVID-R is 5.8 days. Lastly, $R_0 = \beta/\gamma$, since we know $\gamma$ and according to \cite{imai2020report}, $R_0$ is 2.6, the value for $\beta$ is $0.4482$.
\item The initial population of the Susceptible group has been set to 126 million. The initial number of Exposed group has been set to 1000, and the number of the Infectious and Recovered groups has been set to zero. 
\item The number of beds has been set to 1.5 per 1000 inhabitants and we estimate that 5\% of the people contracting the virus need to be sent to ICUs in hospitals \cite{lai2020severe,he2020coronavirus}.
\item For the sake of simplification, we would consider that the population remains constant (not altered by demographics) and that the borders are closed (no immigration or emigration).
\end{itemize}

As explained in section \ref{model-formulation}, to carry out the experiments we defined a new parameter called $\delta$ that can be understood as a weight between 0 and 1 that diminishes the contact rate $\beta$. Therefore we applied the already defined Equations (\ref{eq:op_S}, \ref{eq:op_E}, \ref{eq:op_I}), and (\ref{eq:op_R}) for simulating the evolution of the disease.

\subsubsection{DQL configuration}
For implementing DQL, we use an array to represent a window size of 25 points (where each point represents the percentage of Infectious population for one of the last 25 days), we set a decay rate of 0.99, and a minimum epsilon rate of 0.02 and run the model for 1,000 episodes. The agent has been implemented $\gamma$ = 0.95, with a memory buffer size for 10,000 actions. Every ten days we test the model to see its performance and evaluate the total rewards. To approximate the Q-Table we used a Deep Learning Network based in a densely connected sequential multi-layer perceptron \cite{denoyer2014deep} with four layers with 64, 128, 128, and 8 nodes respectively. The ANN utilizes a linear function and an Adam optimizer with 0.001 with mean-squared error as the loss function. 

\subsubsection{GA configuration}
For the GA, we use a population of 100 individuals (the standard size). We have used a mutation type "random" with a probability of 60\%. We also used a crossover type "uniform" with a probability of 60\%. We kept the best parent and the parent selection was implemented with the "tournament" type with $k=3$, where $k$ represents the number of individuals in the tournament. The number of generations was set to 1000. The crossover was "uniform" and the mutation type "random" \cite{back1994selective}. The percentage of mutation in the genes was set to 10\%.

\subsection{Experiment I: Keeping the ICU below the saturation point}

For conducting the first experiment we used the scenario described in section \ref{problem}, and we implemented a GA and a DQL approach. In order to have a baseline, we generated 1000 different sequences of actions randomly and selected the one with the highest performance. The results of Table \ref{tabla-exp1} show the average, maximum value, minimum value, and standard deviation of 100 executions with GA, DQL, and random actions.

\begin{table}[htbp]
\centering
\caption{Results for RL, GA, and Random actions obtained in the Experiment I.}
\begin{tabular}{|c|c|c|c|c|c|}
\hline
 & \textbf{Avg} & \textbf{Max} &\textbf{ Min} & \textbf{Std} & \textbf{Time (Sec)} \\ \hline
DQL &  1658.06  &   1734  &   1492  &   54.93  &   284.2 \\ \hline
GA &  1648.19  &   1676  &   1620  &   11.74  &   164.68 \\ \hline
Random &  1487.76  &   1504  &   1476  &   5.88  & --- \\ \hline
\end{tabular}
\label{tabla-exp1}
\end{table}

\begin{figure}[t]
\centering
\includegraphics[width=0.50\linewidth]{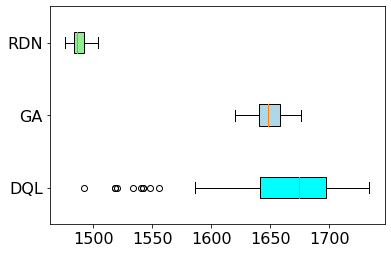}
\caption{Average performance in Experiment I for 100 executions of DQL, GA, and 1000 randomly generated sequences.}
\label{fig:Exp1.png}
\end{figure}

The results can also be seen in a more graphical way in Figure \ref{fig:Exp1.png}. This shows how DQL average is better than that of GA but also that the performance varies a lot for each execution.

\begin{figure}[t]
\centering
\includegraphics[width=0.5\linewidth]{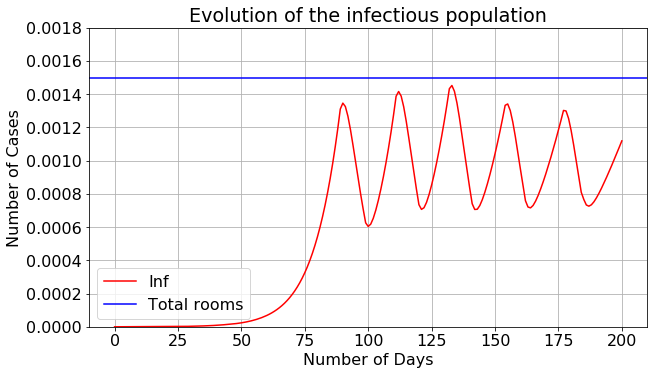}
\caption{Evolution of one of the best experiments for one of the best sequences of actions using DQL.}
\label{fig:Exp1-DQL}
\end{figure}

Additionally, we also displayed the results of the best sequence with DQL in Figure \ref{fig:Exp1-DQL}, which got a total reward of 1734 billion. As can be seen, the number of people (represented in red colour) requiring ICU never surpasses the number of available beds in the hospitals (the blue line).

\begin{figure}[t]
\centering
\includegraphics[width=0.50\linewidth]{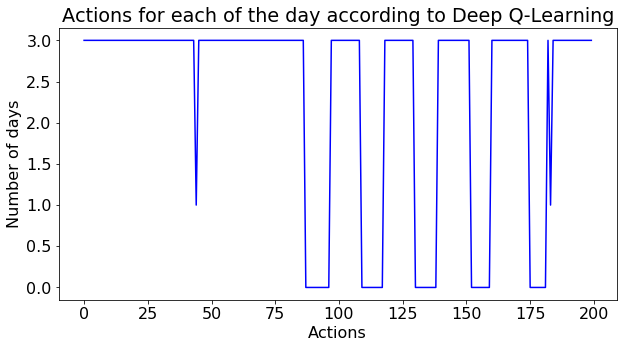}
\caption{DQL recomended actions during the 200 days for the sequence with highest rewards.}
\label{fig:Exp1-Actions-DQL}
\end{figure}

As can be seen in Figure \ref{fig:Exp1-Actions-DQL}, which is the sequence with the highest total rewards, the DQL model mostly recommends  combinations of total lock-downs (phase 0) or no restrictions at all (phase 3). 

\begin{figure}[t]
\centering
\includegraphics[width=0.5\linewidth]{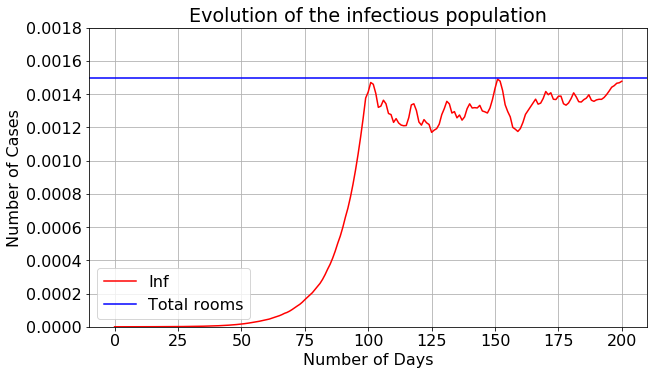}
\caption{Evolution of one of the best experiments for one of the best sequences of actions.}
\label{fig:Exp1-GA}
\end{figure}

Likewise, we also display the best sequence obtained with GA in Figure \ref{fig:Exp1-GA}.

\begin{figure}[t]
\centering
\includegraphics[width=0.5\linewidth]{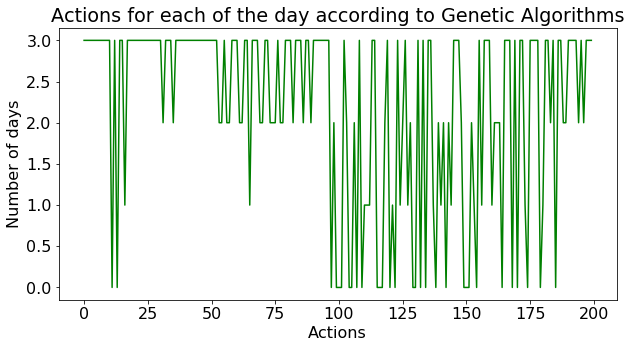}
\caption{Evolution of one of the best experiments for one of the best sequences of actions using GA.}
\label{fig:Exp1-Actions-GA}
\end{figure}

The recommended actions for the given time interval of 200 days are shown in Figure \ref{fig:Exp1-Actions-GA}.  Unlike DQL, GA gives actions (in green colour) much more dispersed where 15\% belong to phase 0, $8\%$ to phase 1, $18\%$ to phase 2, and lastly, 58\% to phase 3.

\subsection{Experiment II: Adapting to the ICU and temperature fluctuation}

The second experiment is conducted with the same conditions that the first one but with two new conditions: 
\begin{itemize}
\item The number of beds decreases gradually from 1.5 per 1000 inhabitants to 0.5. 
\item The temperature and humidity represented with $\theta$ rises so that the probability of contagion decreases as happens with other viruses like Influenza \cite{lowen2007influenza}. To represent the impact of temperature on the contact rate, the $\theta$ value starts at 1 and it will gradually decreases to 0.5 during the 200 days.
\end{itemize}

The rest of the equations remain constant. We regulate these new equations, in which the contact is reduced not only by $\delta$ (due to the confinement) but also by $\theta$ (due to higher humidity and temperature). Equation (\ref{eq:op_S2}) is now used for calculating the Susceptible group, and equation (\ref{eq:op_E2}) is used to calculate the Exposed group. The other equations do not vary.

\begin{equation}\label{eq:op_S2}
S_i= S_{i-1} - \theta_{i}\delta_{i}\beta S_{i-1} I_{i-1}
\end{equation}

\begin{equation}\label{eq:op_E2}
E_i= E_{i-1} + (\theta_{i}\delta_{i} \beta S_{i-1} I_{i-1} -\alpha E_{i-1})
\end{equation}

For the second experiment, as shown in Table \ref{tabla-exp2}, we calculated the average, the maximum value, the minimum value, and the standard deviation of 100 executions with each of the methods.

\begin{table}[htbp]
\centering
\caption{Average performance in Experiment II for 100 executions of DQL, GA, and 1000 randomly generated sequences.}
\begin{tabular}{|c|c|c|c|c|c|}
\hline
 & \textbf{Avg} & \textbf{Max} & \textbf{Min} & \textbf{Std} & \textbf{Time (Sec)} \\ \hline
DQL &  1830.9   &   1856   &   1768  &   21.9  &   233.52 \\ \hline
GA &  1769.4  &   1794  &   1748  &   10.47  &   127.91 \\ \hline
Random &  1501.91  &   1532  &   1476  &   10.74  & --- \\ \hline
\end{tabular}
\label{tabla-exp2}
\end{table}

\begin{figure}[t]
\centering
\includegraphics[width=0.50\linewidth]{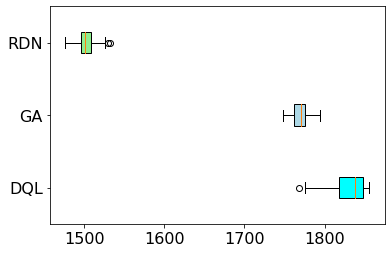}
\caption{For Experiment II, the approach based on DQL is higher than that of GA.}
\label{fig:Exp2.png}
\end{figure}

\begin{figure}[t]
\centering
\includegraphics[width=0.5\linewidth]{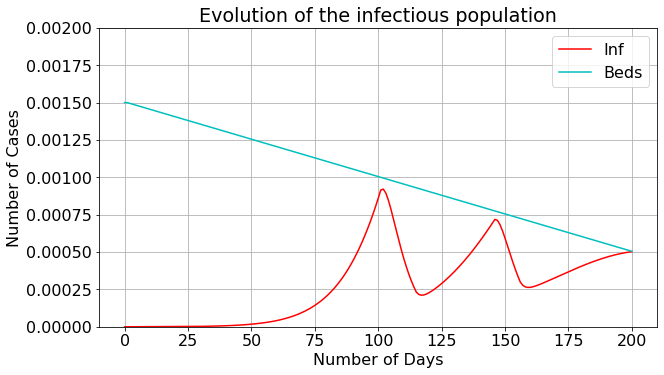}
\caption{Evolution of one of the best experiments for one of the best sequences of actions using DQL.}
\label{fig:Exp2-DQL.png}
\end{figure}

The results can be better seen in Figure \ref{fig:Exp2.png}, and in Figure \ref{fig:Exp2-DQL.png} it is displayed the best sequence of actions which has a total summation of rewards of 1856 billion where the maximum number was 200 billion. In red colour, the number of patients that require beds never surpasses the available number of beds. 

Figure \ref{fig:Exp2.png} shows that DQL (in blue colour) has a higher average, maximum number, and minimum number than GA. And also that the variation of the experiments is much lower than in the first experiment even though the parameters of the DQL remained constant.

Figure \ref{fig:Exp2-DQL.png} is showing that the percentage of the infectious group in the population is never higher than the number of available beds. Otherwise, the sequence would have obtained a penalisation and would not be the highest one. So the best solutions are near the line but never go higher. As can be seen, the cost of not trying to harm the economy with restrictive actions is that the number of the infectious population tends to grow.

\begin{figure}[t]
\centering
\includegraphics[width=0.50\linewidth]{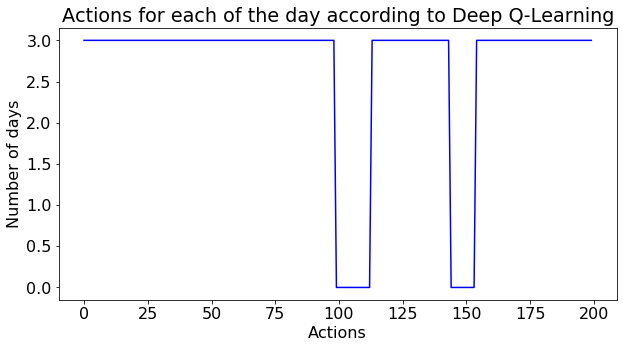}
\caption{For Experiment II, the results of DQL are much better than those of GA.}
\label{fig:Exp2-Actions-DQL}
\end{figure}

In Figure \ref{fig:Exp2-Actions-DQL} we can see the sequence of actions that maximize the total rewards. The DQL model discovers by itself that it is better to coordinate actions of total restrictions (phase 0) or no restrictions at all (phase 3). That is according to the initial reward system. If we had trained the model on a different reward system, the sequences could be very different. For this experiment onwards we do not consider necessary showing the best sequences of the GA because its performance is lower than the one of DQL.

\subsection{Experiment III: Limiting the number of changes in the actions}

For this third experiment, we kept the condition of experiment II of diminishing the number of beds gradually from 1.5 to 0.5 per 1000 inhabitants but we did not consider the condition of the temperature. One of the problems that we see in experiment II is that the kind of actions can change from one day to the next which is not good for citizens. The best thing would be to give citizens the minimum number of changes between phases in order to avoid disorienting them. For this reason, in this third experiment, we included a new penalization to avoid sequences of actions that change from one to other phase all the time.

The penalty for changing for kind of phase to another in consecutive days is not as strong as in the case where the number of patients exceeds the number of beds. On the top of that, we want actions to go from more restrictive to less restrictive so people can go back to normality bit by bit. To solve this problem we use regular expressions. Specifically, we implement the formula pattern =  $``\;\hat{}\;0^*1^*2^*3^*0^* \$"$ with regular expressions that only accepts the following chains. Chains of zeros, followed by chains of ones, threes, and zeros, where each chain has a size of zero or more. Thus, we add the following restriction: "If a day does not fit the pattern of the regular expression, then the reward for that day onward goes to -10". Additionally, since finding sequences of actions in groups is more difficult for the optimization algorithms, we will increase the number of episodes for the DQL and the number of generations in the GA to 2000 instead of 1000 as it was in the first two experiments.

\begin{table}[htbp]
\centering
\caption{Average performance in Experiment III for 100 executions of DQL, GA, and Random.}
\begin{tabular}{|c|c|c|c|c|c|}
\hline
 & \textbf{Avg} & \textbf{Max} & \textbf{Min} & \textbf{Std} & \textbf{Time (Sec)} \\ \hline
DQL &  1568.11   &   1688   &   1378  &   81.86  &   485.26 \\ \hline
GA &  -861.11  &   -736  &   -996  &   61.13  &   353.7 \\ \hline
Random &  -1823.69  &   -1768  &   -1866  &   20.42  & --- \\ \hline
\end{tabular}
\label{tabla-exp3}
\end{table}

\begin{figure}[t]
\centering
\includegraphics[width=0.50\linewidth]{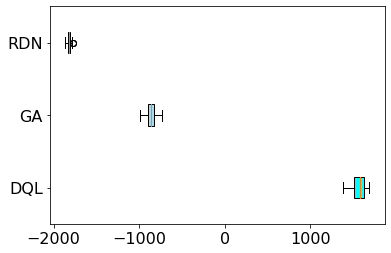}
\caption{For Experiment III, the results of DQL are much better than those of GA.}
\label{fig:Exp3}
\end{figure}

As can be seen in Figure \ref{fig:Exp3}, as the degree of difficulty increases, the higher is the difference between DQL and GA. And it is quite surprising that the variance for the DQL is very low compared to the previous experiments and also, the difference between the DQL and the GA increases as the more restrictions come into play.

\begin{figure}[t]
\centering
\includegraphics[width=0.5\linewidth]{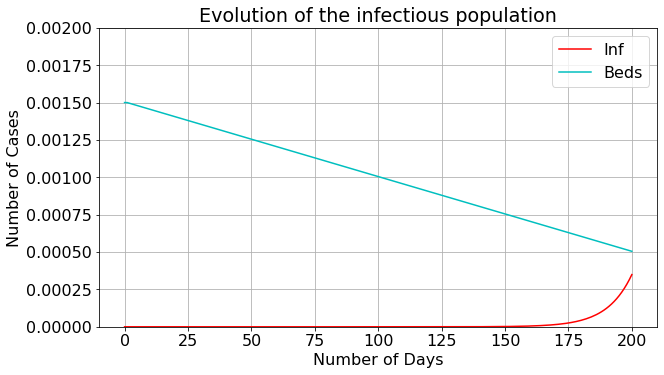}
\caption{Evolution of the actions during days of the best experiments for the best sequences of actions with DQL.}
\label{fig:Exp3-DQL}
\end{figure}

\begin{figure}[t]
\centering
\includegraphics[width=0.50\linewidth]{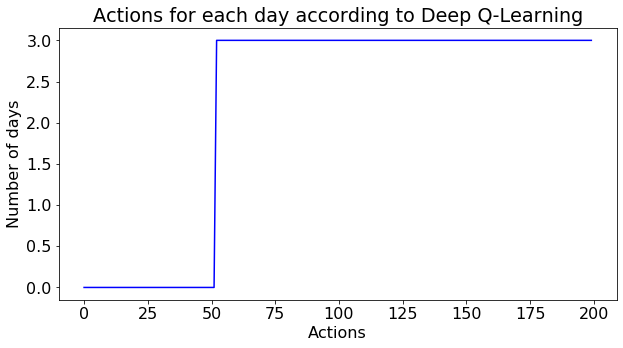}
\caption{The selected actions by the DQL algorithm have very few changes (only one) which was what we were looking for with the new penalization.}
\label{fig:Exp3-Actions-DQL}
\end{figure}

As can be seen in Figure \ref{fig:Exp3-DQL}, the DQL model keeps the infectious population at very low numbers until the moment when it decides not to apply restrictions, liberating the economy at the cost of more virus being more spread in the population. As can be seen in Figure \ref{fig:Exp3-Actions-DQL}, the sequences of actions of the best solution has very few changes between the phases.

\subsection{Discussion}

According to the conducted experiments, the performance of DQL is better than GA and better than the random experiments in all three experiments. In the first experiment, the DQL networks were a bit unstable and the performance varied very much during the executions. Although the best result for the three experiments was always obtained with DQL. 

DQL optimizes this problem better than GA probably because it is able to understand the relationship of one state with the other states thanks to the Bellman's equation. DQL is able to approximate the maximum expected future rewards for each action, according to its previous experience, and consequently, taking the best action in each state.  DQL has a DL neural network that gives you information about the consequences of taking every action in every state. On the other hand, the GA advances by pure combinations but without seeing the big picture and the logic behind the actions. This makes GA to fall very easily into local minimums.

\section{Conclusion and future works}

In this paper, we have simulated the spread of the COVID-19 pandemic over a population using a SEIR epidemiological model. The experiments show that the proposed methodology is a suitable tool for optimizing the governmental actions for a COVID-19 pandemic but it can also be easily extrapolated to other infectious diseases.
Our main contribution has been proposing a methodology, which can be implemented either with GA or DQL, to optimize the actions a government can take to safeguard public health without jeopardising the economy. According to the experiments, the DQL approach provides better solutions in the three experiments. Moreover, the more conditions are included in the reward system the higher is the performance of DQL compared to that of GA. 

As future work, we can also try other approaches such as Policy Gradient which requires less computational resources. Additionally, other optimization methods such as swarm particles can be implemented to see how well they perform. On the other hand, it would be very interesting to adapt our model to a specific country. To this end,  a deeper study of the consequences of a pandemic from an economic point of view is needed.

Finally, there are different methodologies that can be used to find out the best parameters to improve the performance of the different algorithms. Some of the well-known algorithms for that purpose are random search, Bayesian optimization, or BOHB.

\bibliographystyle{unsrt}  
\bibliography{references} 

\begin{thebibliography}{10}

\bibitem{Covidpandemic}
World~Health Organization.
\newblock Coronavirus disease (covid-19) pandemic.

\bibitem{lai2020severe}
Chih-Cheng Lai, Tzu-Ping Shih, Wen-Chien Ko, Hung-Jen Tang, and Po-Ren Hsueh.
\newblock Severe acute respiratory syndrome coronavirus 2 (sars-cov-2) and
  corona virus disease-2019 (covid-19): the epidemic and the challenges.
\newblock {\em International journal of antimicrobial agents}, page 105924,
  2020.

\bibitem{fernandes2020economic}
Nuno Fernandes.
\newblock Economic effects of coronavirus outbreak (covid-19) on the world
  economy.
\newblock {\em Available at SSRN 3557504}, 2020.

\bibitem{he2020coronavirus}
Feng He, Yu~Deng, and Weina Li.
\newblock Coronavirus disease 2019: What we know?
\newblock {\em Journal of medical virology}, 2020.

\bibitem{world2020rational}
World~Health Organization et~al.
\newblock Rational use of personal protective equipment for coronavirus disease
  (covid-19): interim guidance, 27 february 2020.
\newblock Technical report, World Health Organization, 2020.

\bibitem{wearing2005appropriate}
Helen~J Wearing, Pejman Rohani, and Matt~J Keeling.
\newblock Appropriate models for the management of infectious diseases.
\newblock {\em PLoS medicine}, 2(7), 2005.

\bibitem{barro2020coronavirus}
Robert~J Barro, Jos{\'e}~F Urs{\'u}a, and Joanna Weng.
\newblock The coronavirus and the great influenza pandemic: Lessons from the
  “spanish flu” for the coronavirus’s potential effects on mortality and
  economic activity.
\newblock Technical report, National Bureau of Economic Research, 2020.

\bibitem{puterman2014markov}
Martin~L Puterman.
\newblock {\em Markov decision processes: discrete stochastic dynamic
  programming}.
\newblock John Wiley \& Sons, 2014.

\bibitem{hester2018deep}
Todd Hester, Matej Vecerik, Olivier Pietquin, Marc Lanctot, Tom Schaul, Bilal
  Piot, Dan Horgan, John Quan, Andrew Sendonaris, Ian Osband, et~al.
\newblock Deep q-learning from demonstrations.
\newblock In {\em Thirty-Second AAAI Conference on Artificial Intelligence},
  2018.

\bibitem{whitley1994genetic}
Darrell Whitley.
\newblock A genetic algorithm tutorial.
\newblock {\em Statistics and computing}, 4(2):65--85, 1994.

\bibitem{d2002stability}
Alberto d'Onofrio.
\newblock Stability properties of pulse vaccination strategy in seir epidemic
  model.
\newblock {\em Mathematical biosciences}, 179(1):57--72, 2002.

\bibitem{wang2016does}
Fei-Yue Wang, Jun~Jason Zhang, Xinhu Zheng, Xiao Wang, Yong Yuan, Xiaoxiao Dai,
  Jie Zhang, and Liuqing Yang.
\newblock Where does alphago go: From church-turing thesis to alphago thesis
  and beyond.
\newblock {\em IEEE/CAA Journal of Automatica Sinica}, 3(2):113--120, 2016.

\bibitem{kermack1927contribution}
William~Ogilvy Kermack and Anderson~G McKendrick.
\newblock A contribution to the mathematical theory of epidemics.
\newblock {\em Proceedings of the royal society of london. Series A, Containing
  papers of a mathematical and physical character}, 115(772):700--721, 1927.

\bibitem{li1995global}
Michael~Y Li and James~S Muldowney.
\newblock Global stability for the seir model in epidemiology.
\newblock {\em Mathematical biosciences}, 125(2):155--164, 1995.

\bibitem{mnih2013playing}
Volodymyr Mnih, Koray Kavukcuoglu, David Silver, Alex Graves, Ioannis
  Antonoglou, Daan Wierstra, and Martin Riedmiller.
\newblock Playing atari with deep reinforcement learning.
\newblock {\em arXiv preprint arXiv:1312.5602}, 2013.

\bibitem{alimadadi2020artificial}
Ahmad Alimadadi, Sachin Aryal, Ishan Manandhar, Patricia~B Munroe, Bina Joe,
  and Xi~Cheng.
\newblock Artificial intelligence and machine learning to fight covid-19, 2020.

\bibitem{naude2020artificial}
Wim Naud{\'e}.
\newblock Artificial intelligence against covid-19: An early review.
\newblock 2020.

\bibitem{randhawa2020machine}
Gurjit~S Randhawa, Maximillian~PM Soltysiak, Hadi El~Roz, Camila~PE de~Souza,
  Kathleen~A Hill, and Lila Kari.
\newblock Machine learning using intrinsic genomic signatures for rapid
  classification of novel pathogens: Covid-19 case study.
\newblock {\em Plos one}, 15(4):e0232391, 2020.

\bibitem{metsky2020crispr}
Hayden~C Metsky, Catherine~A Freije, Tinna-Solveig~F Kosoko-Thoroddsen,
  Pardis~C Sabeti, and Cameron Myhrvold.
\newblock Crispr-based covid-19 surveillance using a genomically-comprehensive
  machine learning approach.
\newblock {\em bioRxiv}, 2020.

\bibitem{yan2020prediction}
Li~Yan, Hai-Tao Zhang, Yang Xiao, Maolin Wang, Chuan Sun, Jing Liang, Shusheng
  Li, Mingyang Zhang, Yuqi Guo, Ying Xiao, et~al.
\newblock Prediction of survival for severe covid-19 patients with three
  clinical features: development of a machine learning-based prognostic model
  with clinical data in wuhan.
\newblock {\em medRxiv}, 2020.

\bibitem{ge2020data}
Yiyue Ge, Tingzhong Tian, Sulin Huang, Fangping Wan, Jingxin Li, Shuya Li, Hui
  Yang, Lixiang Hong, Nian Wu, Enming Yuan, et~al.
\newblock A data-driven drug repositioning framework discovered a potential
  therapeutic agent targeting covid-19.
\newblock {\em bioRxiv}, 2020.

\bibitem{wang2020covid}
Linda Wang and Alexander Wong.
\newblock Covid-net: A tailored deep convolutional neural network design for
  detection of covid-19 cases from chest radiography images.
\newblock {\em arXiv preprint arXiv:2003.09871}, 2020.

\bibitem{li2020artificial}
Lin Li, Lixin Qin, Zeguo Xu, Youbing Yin, Xin Wang, Bin Kong, Junjie Bai,
  Yi~Lu, Zhenghan Fang, Qi~Song, et~al.
\newblock Artificial intelligence distinguishes covid-19 from community
  acquired pneumonia on chest ct.
\newblock {\em Radiology}, page 200905, 2020.

\bibitem{gozes2020rapid}
Ophir Gozes, Maayan Frid-Adar, Hayit Greenspan, Patrick~D Browning, Huangqi
  Zhang, Wenbin Ji, Adam Bernheim, and Eliot Siegel.
\newblock Rapid ai development cycle for the coronavirus (covid-19) pandemic:
  Initial results for automated detection \& patient monitoring using deep
  learning ct image analysis.
\newblock {\em arXiv preprint arXiv:2003.05037}, 2020.

\bibitem{rao2020identification}
Arni SR~Srinivasa Rao and Jose~A Vazquez.
\newblock Identification of covid-19 can be quicker through artificial
  intelligence framework using a mobile phone--based survey when cities and
  towns are under quarantine.
\newblock {\em Infection Control \& Hospital Epidemiology}, pages 1--5, 2020.

\bibitem{barstugan2020coronavirus}
Mucahid Barstugan, Umut Ozkaya, and Saban Ozturk.
\newblock Coronavirus (covid-19) classification using ct images by machine
  learning methods.
\newblock {\em arXiv preprint arXiv:2003.09424}, 2020.

\bibitem{hu2020artificial}
Zixin Hu, Qiyang Ge, Li~Jin, and Momiao Xiong.
\newblock Artificial intelligence forecasting of covid-19 in china.
\newblock {\em arXiv preprint arXiv:2002.07112}, 2020.

\bibitem{al2020optimization}
Mohammed~AA Al-qaness, Ahmed~A Ewees, Hong Fan, and Mohamed Abd El~Aziz.
\newblock Optimization method for forecasting confirmed cases of covid-19 in
  china.
\newblock {\em Journal of Clinical Medicine}, 9(3):674, 2020.

\bibitem{ardabili2020covid}
Sina~F Ardabili, Amir Mosavi, Pedram Ghamisi, Filip Ferdinand, Annamaria~R
  Varkonyi-Koczy, Uwe Reuter, Timon Rabczuk, and Peter~M Atkinson.
\newblock Covid-19 outbreak prediction with machine learning.
\newblock {\em Available at SSRN 3580188}, 2020.

\bibitem{jahanbin2020using}
Kia Jahanbin and Vahid Rahmanian.
\newblock Using twitter and web news mining to predict covid-19 outbreak.
\newblock {\em Asian Pacific Journal of Tropical Medicine}, 13, 2020.

\bibitem{hoertel2020facing}
Nicolas Hoertel, Martin Blachier, Carlos Blanco, Mark Olfson, Marc Massetti,
  Frederic Limosin, and Henri Leleu.
\newblock Facing the covid-19 epidemic in nyc: a stochastic agent-based model
  of various intervention strategies.
\newblock {\em medRxiv}, 2020.

\bibitem{yang2020modified}
Zifeng Yang, Zhiqi Zeng, Ke~Wang, Sook-San Wong, Wenhua Liang, Mark Zanin, Peng
  Liu, Xudong Cao, Zhongqiang Gao, Zhitong Mai, et~al.
\newblock Modified seir and ai prediction of the epidemics trend of covid-19 in
  china under public health interventions.
\newblock {\em Journal of Thoracic Disease}, 12(3):165, 2020.

\bibitem{libin2020deep}
Pieter Libin, Arno Moonens, Timothy Verstraeten, Fabian Perez-Sanjines, Niel
  Hens, Philippe Lemey, and Ann Now{\'e}.
\newblock Deep reinforcement learning for large-scale epidemic control.
\newblock {\em arXiv preprint arXiv:2003.13676}, 2020.

\bibitem{khadilkar2020optimising}
Harshad Khadilkar, Tanuja Ganu, and Deva~P Seetharam.
\newblock Optimising lockdown policies for epidemic control using reinforcement
  learning.
\newblock {\em arXiv preprint arXiv:2003.14093}, 2020.

\bibitem{liu2020microscopic}
Changliu Liu.
\newblock A microscopic epidemic model and pandemic prediction using
  multi-agent reinforcement learning.
\newblock {\em arXiv preprint arXiv:2004.12959}, 2020.

\bibitem{hausknecht2015deep}
Matthew Hausknecht and Peter Stone.
\newblock Deep recurrent q-learning for partially observable mdps.
\newblock In {\em 2015 AAAI Fall Symposium Series}, 2015.

\bibitem{watkins1992q}
Christopher~JCH Watkins and Peter Dayan.
\newblock Q-learning.
\newblock {\em Machine learning}, 8(3-4):279--292, 1992.

\bibitem{chen2020mathematical}
Tian-Mu Chen, Jia Rui, Qiu-Peng Wang, Ze-Yu Zhao, Jing-An Cui, and Ling Yin.
\newblock A mathematical model for simulating the phase-based transmissibility
  of a novel coronavirus.
\newblock {\em Infectious Diseases of Poverty}, 9(1):1--8, 2020.

\bibitem{imai2020report}
Natsuko Imai, Anne Cori, Ilaria Dorigatti, Marc Baguelin, Christl~A Donnelly,
  Steven Riley, and Neil~M Ferguson.
\newblock Report 3: transmissibility of 2019-ncov.
\newblock In {\em Imperial College London}. 2020.

\bibitem{denoyer2014deep}
Ludovic Denoyer and Patrick Gallinari.
\newblock Deep sequential neural network, 2014.

\bibitem{back1994selective}
Thomas Back.
\newblock Selective pressure in evolutionary algorithms: A characterization of
  selection mechanisms.
\newblock In {\em Proceedings of the first IEEE conference on evolutionary
  computation. IEEE World Congress on Computational Intelligence}, pages
  57--62. IEEE, 1994.

\bibitem{lowen2007influenza}
Anice~C Lowen, Samira Mubareka, John Steel, and Peter Palese.
\newblock Influenza virus transmission is dependent on relative humidity and
  temperature.
\newblock {\em PLoS Pathog}, 3(10):e151, 2007.

\end{thebibliography}
\end{document}